\documentclass[10pt,twocolumn,letterpaper]{article}

\usepackage[pagenumbers]{cvpr} 

\usepackage{graphicx}
\usepackage{amsmath}
\usepackage{amssymb}
\usepackage{booktabs}
\usepackage{bbding}
\usepackage{pifont}
\newcommand{\xmark}{\ding{55}}%

\usepackage[pagebackref,breaklinks,colorlinks]{hyperref}

\usepackage[capitalize]{cleveref}
\crefname{section}{Sec.}{Secs.}
\Crefname{section}{Section}{Sections}
\Crefname{table}{Table}{Tables}
\crefname{table}{Tab.}{Tabs.}

\begin{document}

\title{RAMM: Retrieval-augmented Biomedical Visual Question Answering \\ with Multi-modal Pre-training}

\author{Zheng Yuan$^{1}$, Qiao Jin$^{12}$, Chuanqi Tan$^{1}$, Zhengyun Zhao$^{2}$, Hongyi Yuan$^{12}$, Fei Huang$^{1}$, Songfang Huang$^{1}$\\
$^{1}$Alibaba Group, $^{2}$Tsinghua University \\
{\tt\small \{yuanzheng.yuanzhen,qiao.jqa,chuanqi.tcq,f.huang,songfang.hsf\}@alibaba-inc.com 
} \\
{\tt\small \{zhengyun21,yuanhy20\}@mails.tsinghua.edu.cn}
}
\maketitle

\begin{abstract}
Vision-and-language multi-modal pretraining and fine-tuning have shown great success in visual question answering (VQA).
Compared to general domain VQA, the performance of biomedical VQA suffers from limited data. 
In this paper, we propose a retrieval-augmented pretrain-and-finetune paradigm named \textbf{RAMM} for biomedical VQA to overcome the data limitation issue. Specifically, we collect a new biomedical dataset named \textbf{PMCPM} which offers patient-based image-text pairs containing diverse patient situations from PubMed. Then, we 
pretrain the biomedical multi-modal model to learn visual and textual representation for image-text pairs and align these representations with image-text contrastive objective (ITC).
Finally, we propose a retrieval-augmented method to better use the limited data. We propose to retrieve similar image-text pairs based on ITC from pretraining datasets and introduce a novel retrieval-attention module to fuse the representation of the image and the question with the retrieved images and texts.
Experiments demonstrate that our retrieval-augmented pretrain-and-finetune paradigm obtains state-of-the-art performance on Med-VQA2019, Med-VQA2021, VQARAD, and SLAKE datasets.
Further analysis shows that the proposed RAMM and PMCPM can enhance biomedical VQA performance compared with previous resources and methods. 
We will open-source our dataset, codes, and pretrained model.
\end{abstract}

\section{Introduction}
\label{sec:intro}

\begin{figure}[t!]
    \centering
    \includegraphics[width=0.98\linewidth]{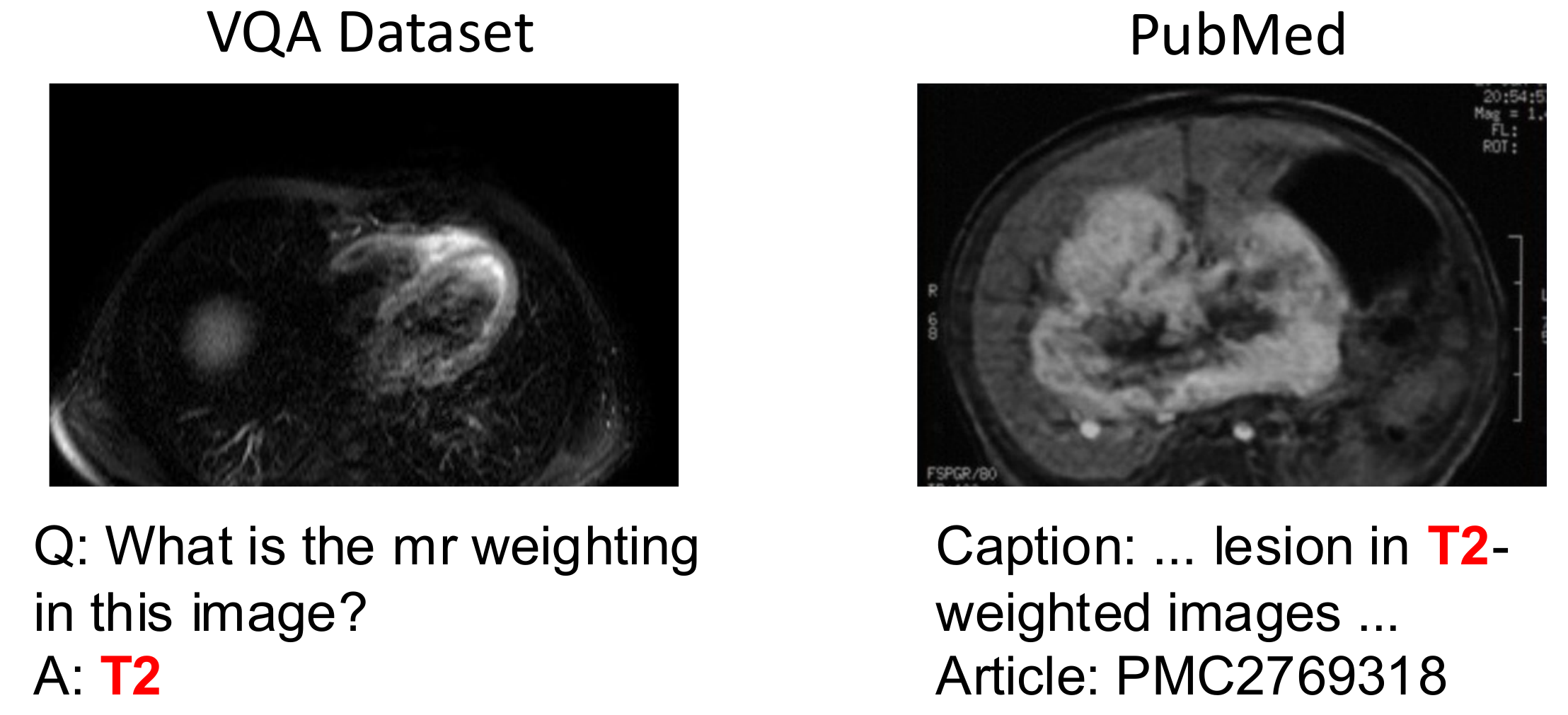}
    \caption{Example of a similar image whose text includes helpful information for answering biomedical questions. The left part is a biomedical-related question with a provided image. The right part is a similar image in PubMed with a relevant caption that indicates the answer (text in red) to the question.
    }
    \label{fig:head}
\end{figure}

Biomedical visual question answering (VQA) aims to answer a biomedical-related question with a provided image.
The images in the biomedical VQA are usually radiology images
(e.g. CT, MRI, and X-Ray), and the questions usually involve modalities, body parts, planes, and abnormal parts in corresponding images.
Answering these questions requires rich knowledge of clinical and medical imaging \cite{bioqa_review}, which is challenging for automated approaches.

With the success of the general domain VQA \cite{albef,dou2022meter,li2022mplug}, some works follow the pretrain-and-finetune paradigm to attempt domain-specific vision-and-language modeling for biomedical VQA  \cite{khare2021mmbert,chen2022m3ae}.
However, unlike the general domain that adequates labeled VQA pairs for fine-tuning, the biomedical domain still suffers from limited data in fine-tuning, which makes models vulnerable to overfitting and hampers the models' ability to learn comprehensive domain knowledge for answering complicated questions.

In contrast to limited data in biomedical VQA datasets, biomedical literature contains abundant and high-quality image-text pairs.
These texts usually describe the images in all aspects including modality, body part, plane, and abnormality which are the information in which biomedical VQA is interested. 
Figure~\ref{fig:head} shows an example that the image-text pair from PubMed\footnote{\url{https://www.ncbi.nlm.nih.gov/pmc/articles/PMC2769318/}} can benefit to answer the question in the VQA dataset. 
Based on this observation, we argue that retrieval is a feasible option for biomedical VQA. This procedure simulates how doctors solve complicated patient cases: doctors search for similar cases to the provided one and propose a solution based on previous experience. 
Therefore, we can retrieve similar image-text pairs from literature to help answer the question. To retrieve related image-text pairs from biomedical literature, there are three key questions to be answered:  \textbf{Where} to retrieve? \textbf{How} to retrieve? and \textbf{How} to leverage the retrieved image-text pairs? 

To this end, we propose a Retrieval-Augmented bioMedical Multi-modal pretrain-and-finetune paradigm (\textbf{RAMM)} for biomedical VQA. Firstly, there are existing biomedical image-text pairs including ROCO \cite{roco} and MIMIC-CXR \cite{mimic-cxr} that can be used for retrieval. However, ROCO is limited by quantity and MIMIC-CXR only focuses on chest X-Ray. A massive, diverse, and high-quality dataset is required for better retrieval. Therefore, we propose PMC-Patients-Multi-modal (\textbf{PMCPM}), a patient-based image-text dataset filtered from PubMed Central, which is larger than ROCO and MIMIC-CXR with various modalities and conditions of images. Secondly, since questions are homogenous in biomedical VQA, we propose to use images for retrieving. We pretrain the biomedical multi-modal model with three tasks: masked language modeling, image-text contrastive  learning, and image-text matching using the combination of our PMCPM dataset and previous widely-used ROCO and MIMIC-CXR datasets.
As the image-text contrastive objective in the pretraining phase aims to align visual and textual representations in the same embedding space \cite{clip,albef}, the image-text contrastive similarity can be used to retrieve similar pairs.
Finally, taking the above datasets as resources, we propose to retrieve the related image-text pair to the given image. To leverage the retrieved pairs, we propose a retrieval-augmented fine-tuning method with a novel retrieval-attention module in the multi-modal encoder layer and apply the fused representation to predict the final answer.

We conduct experiments on four biomedical VQA datasets. Our RAMM achieves state-of-the-art results of 82.13, 39.20, 78.27, and 86.05 scores on VQA-Med 2019 \cite{medvqa2019}, VQA-Med 2021 \cite{medvqa2021}, VQARAD \cite{vqarad}, and SLAKE (English Version) \cite{slake}, respectively. We also conduct ablation studies to verify the effectiveness of our collected PMCPM dataset and the retrieval-augmented fine-tuning methods. Case studies show the retrieved image-text pairs benefit to answer the question.

We summarize our contributions as follows:
\begin{itemize}
    \item We collect PMCPM, a patient-based image-text dataset from PubMed which can be used for biomedical multi-modal pretraining and retrieving.
    \item We propose RAMM, a retrieval-augmented multi-modal pretrain-and-finetune paradigm for biomedical VQA with a novel retrieval-attention module.
    \item Experiments show that the proposed RAMM with additional PMCPM dataset outperforms previous VQA methods on VQA-Med 2019, VQA-Med 2021, VQARAD, and SLAKE datasets. 
\end{itemize}

\section{Related Work}


\paragraph{Biomedical Multi-modal Pretraining} 
In the general domain, multi-modal pretraining models show effectiveness in various image-text-related downstream tasks \cite{albef,dou2022meter,li2022mplug,wang2022image}.
In the biomedical NLP community, previous research has shown that further continue training in the in-domain corpus can enhance the performance of pretrained language models \cite{beltagy-etal-2019-scibert,lee2020biobert,pubmedbert,yuan-etal-2021-improving,yuan-etal-2022-biobart}.
Combining the above-mentioned two points, it is direct to consider pretraining a biomedical in-domain multi-modal model for biomedical VQA.
Previous biomedical pretrained multi-modal models \cite{pubmedclip,khare2021mmbert,medvill,chen2022m3ae,arl} mainly leverage ROCO \cite{roco} and MIMIC-CXR \cite{mimic-cxr} as pretraining datasets.
These models predict masked tokens \cite{chen2022m3ae,arl} or image patches \cite{chen2022m3ae} and conduct image-text alignment \cite{pubmedclip} as pretraining tasks.
Compared with previous models, our RAMM is pretrained with our additional proposed PMCPM dataset which contains images from diverse conditions of patients.
For pretraining tasks, our model adopts masked language modeling, image-text contrastive learning, and image-text matching. Image-text contrastive similarities are used for retrieving in RAMM.

\paragraph{Biomedical Visual Question Answering}
VQA requires answering an image-related question.
Different from general domain VQA \cite{albef,dou2022meter,li2022mplug}, the training samples of biomedical VQA are much fewer.
To overcome the data limitation in biomedical VQA, meta-learning is adapted for quick fitting to VQA questions \cite{nguyen2019overcoming,aioz_mmq_miccai21}.
With the success of the pretrain-and-finetune paradigm, pretraining a multi-modal model with large-scale unsupervised datasets learns better visual and textual representations which mitigate the impact of data limitation and boosts biomedical VQA performances significantly \cite{mevf,pubmedclip,chen2022m3ae}.
In this paper, we also pretrain a biomedical multi-modal model RAMM with patient-based image-text pairs PMCPM.
While fine-tuning VQA datasets, RAMM retrieves similar image-text pairs and tries to leverage useful information from these images and texts.

\paragraph{Retrieval Augmentation} has been applied successfully in NLP \cite{lee2019latent,lewis2020retrieval,guu2020retrieval} and CV \cite{long2022retrieval,sarto2022retrieval} for knowledge-intensive tasks.
In this paper, we decide to use retrieval augmentation to solve the knowledge-intensive and under-labeled biomedical VQA tasks.
A retriever may be sparse like BM-25 \cite{bm25,lee2019latent} or dense using pretrained encoders \cite{guu2020retrieval,sarto2022retrieval}.
For our task, the model pretrained by the image-text contrastive (ITC) task can be naturally used as a multi-modal retriever.
Thus, we pretrain a biomedical multi-modal model by ITC for retrieval augmentation.
We also propose a retrieval-attention module to fuse these retrieved samples.
To the best of our knowledge, this is the first work to leverage retrieval augmentation in biomedical VQA.


\begin{figure*}[t!]
    \centering
    \includegraphics[width=0.8\linewidth]{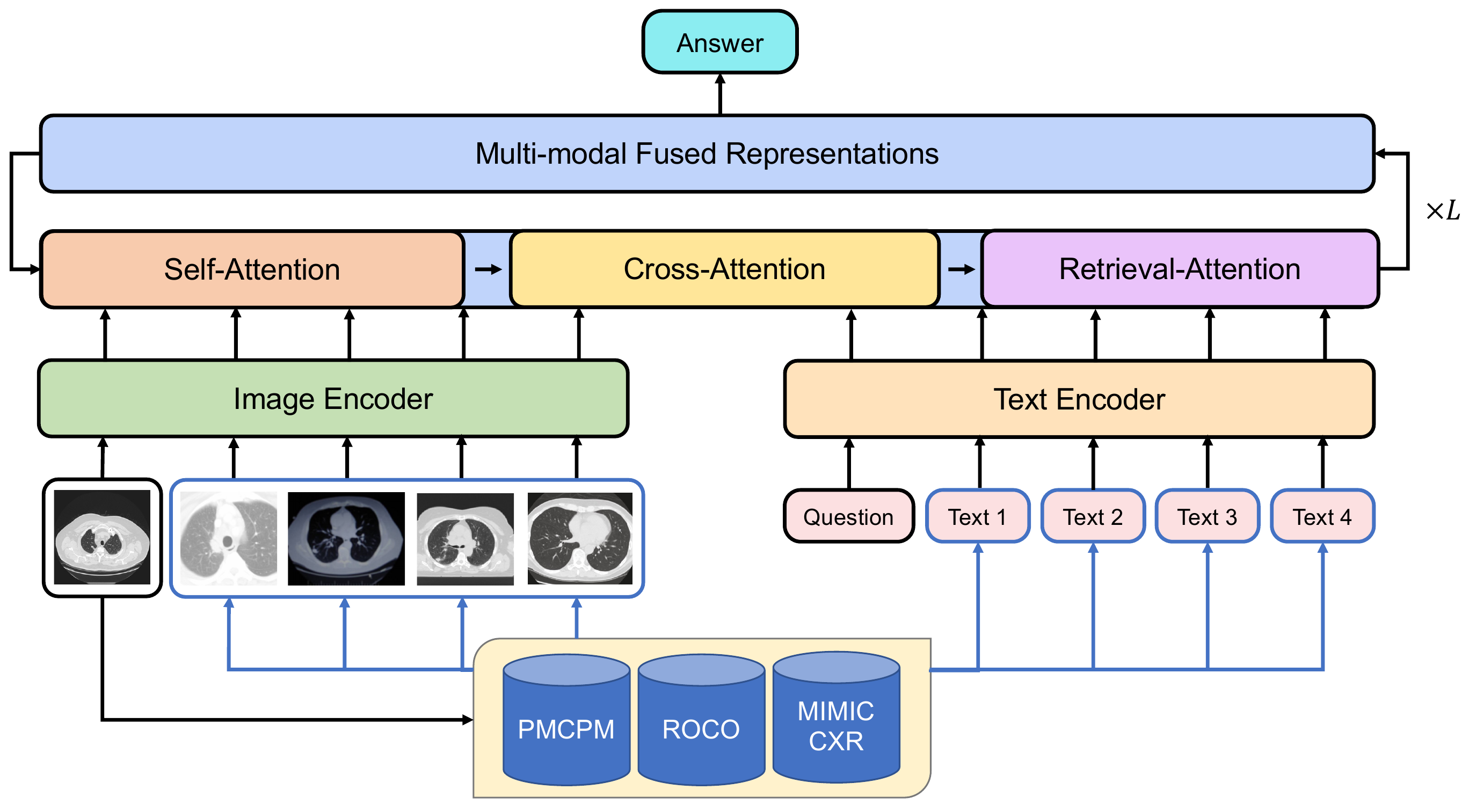}
    \caption{The workflow of our proposed retrieval-augmented paradigm RAMM for biomedical VQA.
    The origin image is from VQA datasets, and the other images are retrieved.
    Blue lines represent retrieving image-text pairs (PMCID:4061445, 2278144, 4725742, and 3616602) from PMCPM, ROCO, and MIMIC-CXR datasets based on image-text contrastive similarities.
    Retrieval-augmented samples are sent into uni-modal and multi-modal encoders. Multi-modal encoders contain retrieval-attention to fuse retrieved information. Representations obtained by multi-modal encoders are used for VQA classification.
    }
    \label{fig:workflow}
\end{figure*}

\section{Approach}

In this section, we first introduce how we build the PMCPM dataset.
We then describe the model architecture and pretraining tasks.
Finally, we introduce how RAMM retrieves image-text pairs from pretraining datasets, and how we combine retrieved samples during biomedical VQA fine-tuning.
Our overall workflow is shown in Figure~\ref{fig:workflow}.

\subsection{PMCPM Dataset}
\label{sec:pmcp}

\begin{figure}[t!]
    \centering
    \includegraphics[width=0.9\linewidth]{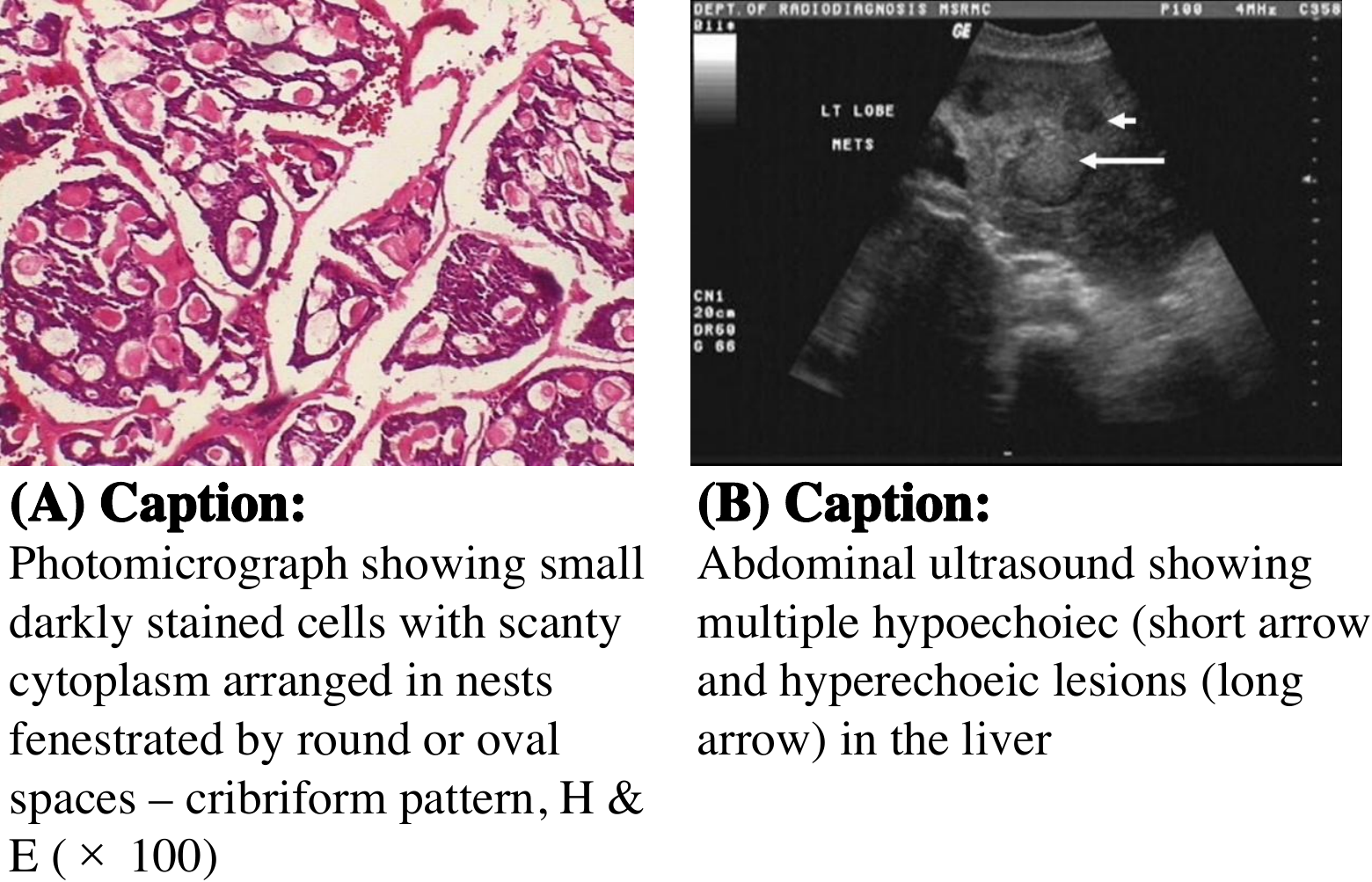}
    \caption{Examples of image-text pairs from PMCPM.
    }
    \label{fig:case}
\end{figure}

To obtain domain-specific image-text pairs of high quality and large scale, we first consider using the Open Access subset of PubMed Central (PMC OA)\footnote{\url{https://www.ncbi.nlm.nih.gov/pmc/tools/openftlist/}}, a free biomedical literature archive of over $4$M publications with both full-text (including figure captions) and figures in these articles publicly available through the official FTP service\footnote{\url{https://www.ncbi.nlm.nih.gov/pmc/tools/ftp/}}. 
However, the figures in PMC OA full set can be quite noisy: apart from clinical images, many kinds of graphs appear in a research article, including flowcharts, histograms, and scatter plots, which provide little information for multi-modal pretraining. 
Pelka \etal\cite{roco} use neural networks trained on human annotated datasets to automatically filter compound, multi-pane, and non-radiology images, collecting a clean dataset with 81k image-text pairs.
To further explore and utilize the figure resource in PMC, we instead resort to a ``patient-based'' collection pipeline inspired by Zhao \etal\cite{zhao2022pmc}.
``\textit{Case Report}'' is a specific type of medical literature, generally describing the whole admission of a patient.
Following Zhao \etal \cite{zhao2022pmc}, we use regular expressions to identify sections containing patient notes, extract the corresponding texts, and apply various filters to remove noises, resulting in a collection of 167k patient notes with high-quality and diverse conditions.
The figures accompanied by these patient notes are closely related to the patient's clinical conditions and contain richer information than solely radiology images. 
We thus collect these figures with corresponding captions to build our PMCPM dataset. Figure~\ref{fig:case} shows two examples of our  PMCPM (PMCID: 509249) with the images and their corresponding captions. Figure~\ref{fig:case}A shows a photomicrograph describing the presence of adenoid cystic carcinoma with cribriform pattern and perineurial invasion. Figure~\ref{fig:case}B shows a radiology image that multiple metastatic lesions scatter in both lobes. The examples show that our patient-based collecting method can target clinical-related figures with more image modality and detailed text descriptions.


\subsection{Model Architecture}

Given an image-text pair $(\it{I},\it{T})$, RAMM first obtains uni-modal features independently and then fuses them for multi-modal representations.
Fused representations are used for VQA classification.
RAMM consists of a text encoder, an image encoder, and a multi-modal encoder for fusing two different modal representations.
The text $\it{T}$ is tokenized into subwords $\{x_{cls}, x_1, ..., x_n\}$ and transformed into hidden representations $\mathbf{w}_0=\{\mathbf{w}_{cls}, \mathbf{w}_1, ..., \mathbf{w}_n\}$ by the text encoder.
The image $\it{I}$ is divided into patches $\{y_{cls}, y_1, ..., y_m\}$ and represented as $\mathbf{v}_0=\{\mathbf{v}_{cls}, \mathbf{v}_1, ..., \mathbf{v}_m\}$ by the image encoder.
The multi-modal encoder is a co-attention style dual-stream transformer.

\paragraph{Multi-modal Fusion Layer}
\label{sec:multi modal fusion layer}
We apply co-attention style multi-modal fusion layers \cite{dou2022meter} upon text features $\mathbf{w}_0$ and visual features $\mathbf{v}_0$ with layer count $L$.
Specifically, co-attention fusion layers consist of dual-stream transformer layers where one stream corresponds to one modal. Each transformer layer is combined with a self-attention module and a cross-attention module\footnote{Feed-forward networks are omitted for notation brevity.}.
For layer $i$, we denote text hidden representations as $\mathbf{w}_i=\{\mathbf{w}_{cls}^i, \mathbf{w}_1^i, ..., \mathbf{w}_n^i\}$ and image hidden representations as $\mathbf{v}_i=\{\mathbf{v}_{cls}^i, \mathbf{v}_1^i, ..., \mathbf{v}_m^i\}$. The attention module in transformers is defined as:
\begin{equation}
    \mathrm{Attn}(Q,K,V) = \mathrm{softmax} (\frac{QK^\top}{\sqrt{d_k}})V
\end{equation}
Using this denotation, self-attention is defined as:
\begin{equation}
    Q^{i,w} = W_Q^{w}\mathbf{w}_i, K^{i,w} = W_K^{w}\mathbf{w}_i, V^{i,w} = W_V^{w}\mathbf{w}_i
\end{equation}
\begin{equation}
    Q^{i,v} = W_Q^{v}\mathbf{v}_i, K^{i,v} = W_K^{v}\mathbf{v}_i, V^{i,v} = W_V^{v}\mathbf{v}_i
\end{equation}
\begin{equation}
    \mathbf{w}'_i = \mathrm{Attn}(Q^{i,w}, K^{i,w}, V^{i,w})
\end{equation}
\begin{equation}
    \mathbf{v}'_i = \mathrm{Attn}(Q^{i,v}, K^{i,v}, V^{i,v})
\end{equation}
We reuse the notation of $Q^{i,\{w,v\}},K^{i,\{w,v\}},V^{i,\{w,v\}}$ by replacing $\mathbf{w}_i$ with $\mathbf{w}'_i$ and $\mathbf{v}_i$ with  $\mathbf{v}'_i$, and cross-attention is defined as:
\begin{equation}
    \mathbf{w}_{i+1} = \mathrm{Attn}(Q^{i,w}, K^{i,v}, V^{i,v})
\end{equation}
\begin{equation}
    \mathbf{v}_{i+1} = \mathrm{Attn}(Q^{i,v}, K^{i,w}, V^{i,w})
\end{equation}
The fused representations of text and image are denoted as $\mathbf{w}_{L}$ and $\mathbf{v}_{L}$.

\subsection{Pretraining Tasks} 
To pretrain RAMM, we employ three pretraining tasks including Image-Text Contrastive learning (ITC) \cite{albef}, Image-Text Matching (ITM) \cite{albef}, and Masked Language Modeling (MLM) \cite{bert}.
The pretraining objective function of RAMM is defined by the sum of these three tasks. ITC aligns uni-modal visual and textual representations $\mathbf{w}=g_w(\mathbf{w}_{cls})$ and $\mathbf{v}=g_v(\mathbf{v}_{cls})$ in the same embedding space, where $g_w$ and $g_v$ are two linear projections with $L^2$ normalization.
ITM predicts whether one image and one text match each other based on fused representations $\mathbf{w}_{L}$ and $\mathbf{v}_{L}$.
MLM predicts masked token in text $\it{T}$ based on fused textual representations $\mathbf{w}_{L}$.
Since ITC aims to align images and texts with cosine similarities, it becomes a natural choice to retrieve related images and texts.


\subsection{Retrieval-augmented Fine-tuning}

We use image-text contrastive similarity to retrieve related image-text pairs and fuse them with a retrieval-attention module in fine-tuning.

\label{section:retrieval}

\paragraph{Retrieve related image text pairs}
Let $\mathcal{D}=\{(\it{I}_j,\it{T}_j)|j\in \mathcal{A}\}$ denotes the set of all image-text pairs, where $\mathcal{A}$ is an index set.
For image-question pair $(\it{I},\it{T})$ from VQA datasets, we retrieve $r$ augmented samples from $\mathcal{D}$ based on uni-modal image representations $\textbf{v}_{cls}$.
Here we do not use the representations from question $\it{T}$ to retrieve since questions in VQA datasets are homogeneous and non-informative. 
With different images, questions can be similar to \textit{What is the abnormality in this image?} or \textit{Is there an abnormality in this image?}
For image-text pair $(\it{I}_j,\it{T}_j) \in \mathcal{D}$ with uni-modal representations $\textbf{v}_{cls}^j$ and $\textbf{w}_{cls}^j$, we calculate the similarities between image $\it{I}$ using ITC similarity scores:
\begin{equation}
    s_w^j = g_v(\mathbf{v}_{cls})^\top g_w(\mathbf{w}^j_{cls})
\end{equation}
\begin{equation}
    s_v^j = g_v(\mathbf{v}_{cls})^\top g_v(\mathbf{v}^j_{cls})
\end{equation}
ITC optimizes $g_v(\mathbf{v}^j_{cls})^\top g_w(\mathbf{w}^j_{cls})\rightarrow 1$ while pretraining, and this leads $s_w^j$ and $s_v^j$ to be close.
So we use the maximum of $s_w^j$ and $s_v^j$ to define the similarity between $(\it{I}_j,\it{T}_j)$ with $\it{I}$:
\begin{equation}
    s_j = \max(s_w^j, s_v^j)
\label{eq:prob}
\end{equation}
We freeze the uni-modal encoders while calculating $s_j$. The first reason is that ITC scores may change dramatically while fine-tuning. The second reason is representations from the pretraining dataset can be pre-computed, and similarity calculation can be conducted by Faiss \cite{faiss} quickly.

We first calculate top-$r$ $\{s_w^j|j\in\mathcal{R}_w\}$ among $\{s_w^j|j\in\mathcal{A}\}$ and top-$r$ $\{s_v^j|j\in\mathcal{R}_v\}$ among $\{s_v^j|j\in\mathcal{A}\}$ using Faiss.
$\mathcal{R}_w \cup \mathcal{R}_v$ contains $[r,2r]$ image-text pairs.
When training, we randomly select $r$ image-text pairs from $\mathcal{R}_w \cup \mathcal{R}_v$.
The probabilities of each image-text pair are rational to similarities $s^j$.
When inference, we select top-$r$ similar image-text pairs from $\mathcal{R}_w \cup \mathcal{R}_v$ based on $s^j$.
We denote retrieved $r$ image-text pairs as $\{(\it{I}_j,\it{T}_j)|1\leq j\leq r\}$.


\paragraph{Retrieval fusion in multi-modal encoder}

\begin{figure}[t!]
    \centering
    \includegraphics[width=0.95\linewidth]{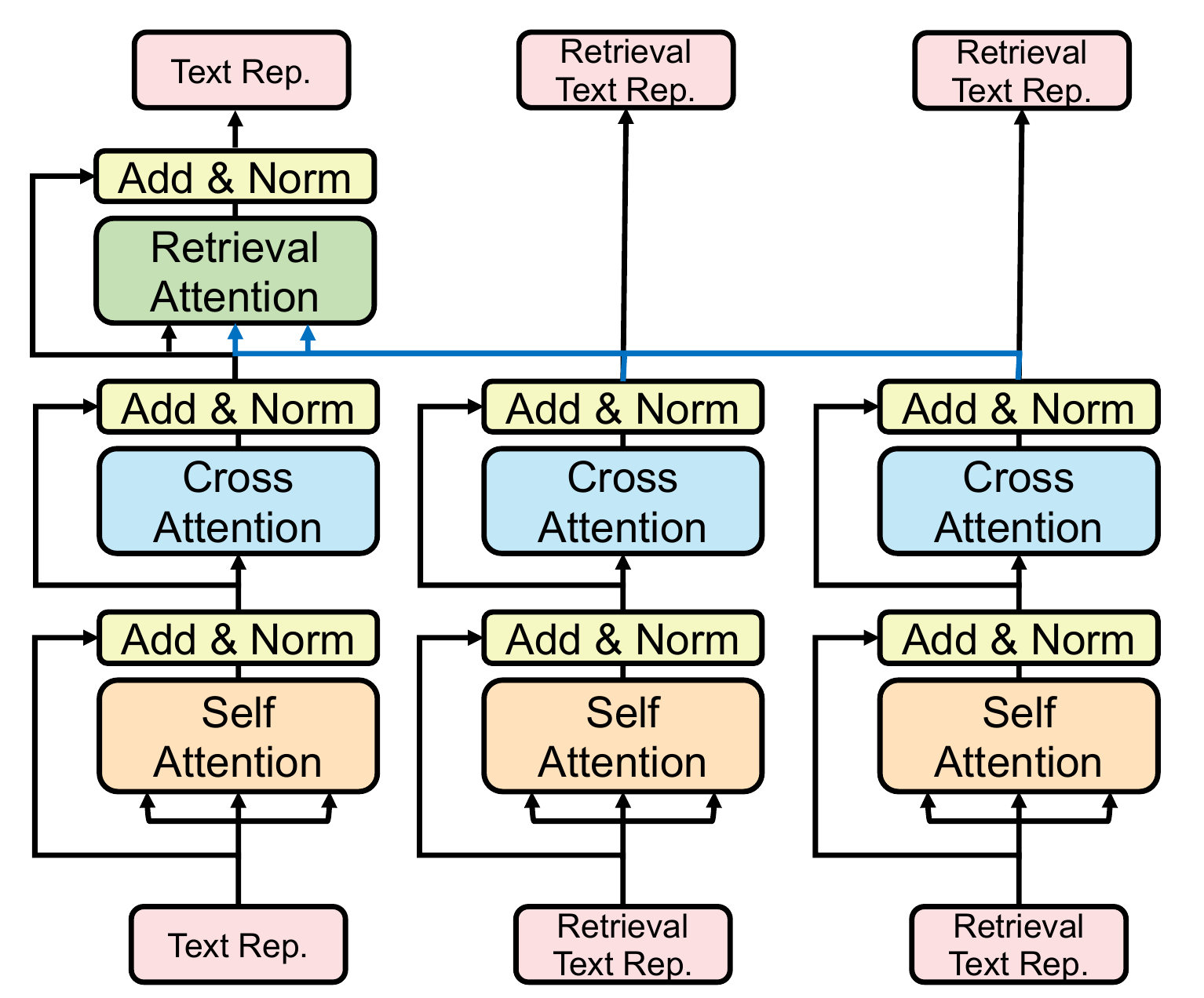}
    \caption{The architecture of retrieval-attention for text modality.
    }
    \label{fig:retrieval-attention}
\end{figure}

To fuse the knowledge of retrieved samples, we propose a retrieval-attention module in the multi-modal fusion layer.
The retrieval-attention module is applied after the self-attention and cross-attention in each multi-modal fusion layer.
Figures~\ref{fig:retrieval-attention} shows the model architecture of retrieval-attention applied to text modality.
Specifically, we use uni-modal encoders encode retrieved image-text pairs $\{(\it{I}_j,\it{T}_j)|1\leq j\leq r\}$ to $\{(\mathbf{w}_0^j,\mathbf{v}_0^j)|1\leq j\leq r\}$. 
We denote the $(\mathbf{w}_0^0,\mathbf{v}_0^0)=(\mathbf{w}_0,\mathbf{v}_0)$ for symbol unification.
For $i^{th}$ multi-modal fusion layer, inputs of text hidden representations are $\{\mathbf{w}_i^j|0\leq j\leq r\}$ and image hidden representations are $\{\mathbf{v}_i^j|0\leq j\leq r\}$.
By applying self-attention and cross-attention defined in Section~\ref{sec:multi modal fusion layer}, we obtain text representations $\{{\mathbf{w}_i^j}'|0\leq j\leq r\}$ and image representations $\{{\mathbf{v}_i^j}'|0\leq j\leq r\}$. 
Retrieval-attention is applied to each modal independently. 
Takes text representations as an example, retrieval-attention uses the $\rm{[CLS]}$ representations of ${\mathbf{w}_i^0}'$ as attention queries and $\rm{[CLS]}$ representations of $\{{\mathbf{w}_i^j}'|0\leq j\leq r\}$ as attention keys and values.
The outputs of retrieval attention are only added to original representations ${\mathbf{w}_i^0}'$ but not retrieved representations $\{{\mathbf{w}_i^j}'|1\leq j\leq r\}$.

\begin{equation}
    Q^{i,r} = W_Q^r{\mathbf{w}_{i,cls}^0}'
\end{equation}
\begin{equation}
    K^{i,r} = W_K^r[{\mathbf{w}_{i,cls}^0}',{\mathbf{w}_{i,cls}^1}',...,{\mathbf{w}_{i,cls}^r}']
\end{equation}
\begin{equation}
    V^{i,r} = W_V^r[{\mathbf{w}_{i,cls}^0}',{\mathbf{w}_{i,cls}^1}',...,{\mathbf{w}_{i,cls}^r}']
\end{equation}
\begin{equation}
     {\mathbf{w}_{i+1,cls}^j} = 
\begin{cases}
     {\mathbf{w}_{i,cls}^j}' + \mathrm{Attn}(Q^{i,r},K^{i,r},V^{i,r}), & j = 0  \\
     {\mathbf{w}_{i,cls}^j}', & j > 0
\end{cases}
\end{equation}

The retrieval-attention on image modality is calculated similarly. 
\begin{equation}
    Q^{i,r} = W_Q^r{\mathbf{v}_{i,cls}^0}'
\end{equation}
\begin{equation}
    K^{i,r} = W_K^r[{\mathbf{v}_{i,cls}^0}',{\mathbf{v}_{i,cls}^1}',...,{\mathbf{v}_{i,cls}^r}']
\end{equation}
\begin{equation}
    V^{i,r} = W_V^r[{\mathbf{v}_{i,cls}^0}',{\mathbf{v}_{i,cls}^1}',...,{\mathbf{v}_{i,cls}^r}']
\end{equation}
\begin{equation}
     {\mathbf{v}_{i+1,cls}^j} = 
\begin{cases}
     {\mathbf{v}_{i,cls}^j}' + \mathrm{Attn}(Q^{i,r},K^{i,r},V^{i,r}), & j = 0  \\
     {\mathbf{v}_{i,cls}^j}', & j > 0
\end{cases}
\end{equation}

We perform retrieval-attention on original representations ${\mathbf{w}_i^0}'$ or ${\mathbf{v}_i^0}'$ to fuse additional information from $\{{\mathbf{w}_i^j}'|1\leq j\leq r\}$ or $\{{\mathbf{v}_i^j}'|1\leq j\leq r\}$.
We do not apply it to retrieved samples since we do not need to exchange information among them.
We use the final layer of origin samples representations $({\mathbf{w}_L^0},{\mathbf{v}_L^0})$ for VQA classification.

\section{Experiments}

\subsection{Dataset}

We conduct experiments on four biomedical VQA datasets as VQA-Med 2019 \cite{medvqa2019}, VQA-Med 2021 \cite{medvqa2021}, VQARAD \cite{vqarad}, and SLAKE (English Version) \cite{slake}. 
To fairly compare with previous works, we follow the same data split with Chen \etal \cite{chen2022m3ae} for VQA-Med 2019, VQARAD, and SLAKE, as well as Gong \etal \cite{gong2021sysu-hcp} for VQA-Med 2021.
We use accuracy to measure model performance.
Detailed statistics of these datasets are listed in Appendix A.

\subsection{Baselines}
We compare RAMM with non-pretrained methods \cite{mfb,san,ban,gong2021sysu-hcp,delbrouck-etal-2022-vilmedic} and pre-trained methods \cite{mevf,cprd,condr} in biomedical VQA.
We introduce the following methods since they are more related to us:

\noindent\textbf{PubmedCLIP} \cite{pubmedclip} fine-tunes CLIP with ROCO to obtain the biomedical image and text representations. 

\noindent\textbf{MMBert} \cite{khare2021mmbert} feeds CNN features of images with masked tokens to apply masked language model pretraining.

\noindent\textbf{MTPT} \cite{mtpt} proposes a multi-task pretraining with an image understanding task and a question-image compatibility task.  

\noindent\textbf{M3AE} \cite{chen2022m3ae} pretrains a biomedical multi-modal model by reconstructing masked tokens and images.


\subsection{Implementation Details}

\paragraph{Pre-training Data}

As introduced in Section~\ref{sec:pmcp}, we collect 398k biomedical image-text pairs of publication quality and various types from PMC OA as PMCPM. Then, we combine the PMCPM with two widely-used biomedical pretrained datasets ROCO \cite{roco} and MIMIC-CXR \cite{mimic-cxr} for model pre-training. Following the previous work, we only use their train splits as pretraining datasets and only use images in frontal view for MIMIC-CXR.
The statistics of training data are shown in Table~\ref{tab:dataset}.

\begin{table}[t]
\centering
\begin{tabular}{lcc}
\toprule
Dataset & Count & Sources \\
\midrule
ROCO & 70,306 & PubMed \\
MIMIC-CXR & 225,829 & BIDMC \\
PMCPM & 398,109  & \textit{Case Report} \\
\bottomrule
\end{tabular}
\caption{Pretraining datasets statistics, where count reports the number of image-text pairs in their train splits. BIDMC is the abbreviation of Beth Israel Deaconess Medical Center.}
\label{tab:dataset}
\end{table}

\paragraph{Pre-training Setting}


For the uni-modal encoders, we initialize the text encoder by PubmedBert \cite{pubmedbert} and the image encoder by Swin-Base \cite{swin}.
For the multi-modal encoder, we apply a 6-layer Transformer \cite{vaswani2017attention} with hidden dimension $d=768$ and head count $n_{head}=12$ with random initialization.
Images are transformed into resolutions $224 \times 224$ with RandAugment \cite{randaugment}.
We apply the momentum distillation \cite{albef} with the momentum parameter as 0.995. 
The model is trained for 30 epochs with a batch size of 512.
We use AdamW \cite{loshchilov2017decoupled} to optimize our models with a linear learning rate decay.
For the detailed pretraining hyper-parameters, please refer to Appendix B.

\paragraph{Fine-tuning Setting}
Following previous works, we regard biomedical VQA as a classification problem and optimize the model by the cross-entropy loss.
For all datasets, we fine-tune our pretrained models with 100 epochs. 
We use frozen uni-modal encoders of RAMM to retrieve image-text pairs as mentioned in Section~\ref{section:retrieval}.
We retrieve $r=4$ pairs for each instance from the combination of ROCO, MIMIC-CXR, and PMCPM.
Images are transformed into resolutions $224 \times 224$ while fine-tuning.
In VQA-Med 2019, VQARAD, and SLAKE, we center crop images, while we use RandAugment \cite{randaugment} in VQA-Med 2021.
Exponential moving average $\theta=0.999$ and R-Drop \cite{liang2021rdrop} $\alpha=0.6$ are also used while fine-tuning.
We list detailed hyper-parameters in Appendix C.

\begin{table}[t]
\small
\centering
\begin{tabular}{lccc}
\toprule
Model & VQAMed2019 & VQARAD & SLAKE \\
\midrule
\emph{No Pretraining} \\
MFB \cite{mfb} & - & 50.60 & 73.30 \\
SAN \cite{san} & - & 54.30 & 76.00 \\
BAN \cite{ban} & - & 58.30 & 76.30 \\
\midrule
\emph{With Pretraining} \\
MEVF-BAN \cite{mevf} &  77.86 & 66.10 & 78.60 \\
CPRD-BAN \cite{cprd} &  - & 67.80 &  81.10 \\
COND-RE \cite{condr} & - & 71.60 & - \\
PubmedCLIP \cite{pubmedclip} & - & 72.10 & 80.10 \\
MMBert \cite{khare2021mmbert} & 77.90 & 72.00 & - \\
MTPT \cite{mtpt} & - & 73.20 & -\\
M3AE \cite{chen2022m3ae} & 79.87 & 77.01 & 83.25 \\
\midrule
\bf{RAMM} & \textbf{82.13} & \textbf{78.27} & \textbf{86.05} \\
w/o retrieval & 80.27 & 76.72 & 85.58 \\
\bottomrule
\end{tabular}
\caption{Results on the VQA-Med 2019, VQARAD, and SLAKE dataset.}
\label{tab:main}
\end{table}

\begin{table}[t]
\small
\centering
\begin{tabular}{lc}
\toprule
Model & VQA-Med 2021 \\
\midrule
VilMedic * \cite{delbrouck-etal-2022-vilmedic} &  37.80 \\
SYSU-HCP * \cite{gong2021sysu-hcp} & 38.20 \\
\bf{RAMM} &  \textbf{39.20} \\
\bottomrule
\end{tabular}
\caption{Results on the VQA-Med 2021. * indicates ensembled results.}
\label{tab:sub}
\end{table}

\subsection{Main Results}
We compare our approach with baseline methods in four datasets.
As shown in Table \ref{tab:main}, methods with pretrained models outperform non-pretrained models by a large margin.
Our proposed RAMM achieves 82.13,  78.27, and 86.05 on VQA-Med 2019,  VQARAD, and SLAKE respectively, which exceeds the previous best result by 1.86, 1.26, and 2.80 respectively. Table 4 shows results on the VQA-Med 2021 dataset. Our RAMM achieves 39.20 in terms of accuracy, which even outperforms the state-of-the-art ensemble methods with 1.00 in the VQA-Med 2021 Challenge.


We also report the result of our RAMM without retrieving additional image-text pairs on the VQA-Med 2019, VQARAD, and SLAKE datasets. As shown in Table \ref{tab:main}, it achieves 80.27, 76.72, and 85.58 on VQA-Med 2019, VQARAD, and SLAKE respectively.
Without retrieval augmentation, RAMM still outperforms the previous state-of-the-art method M3AE on VQA-Med 2019 with 0.4 as well as SLAKE with 2.33 and performs on par on VQARAD (76.72 v.s. 77.01).
This indicates that the pretraining with additional PMCPM dataset learns better visual and textual representations for biomedical VQA tasks.
Compared with models without retrieval augmentation, we observe that our RAMM achieves consistent improvements of 1.90, 1.55, 0.47 on VQA-Med 2019,  VQARAD, and SLAKE respectively, which demonstrates that our proposed retrieval augmentation can boost biomedical VQA performances.


\begin{table*}[t]
\small
\centering
\begin{tabular}{cccccccc}
\toprule
& \multicolumn{2}{c}{Pre-training}  & \multicolumn{2}{c}{Retrieval}  & \multicolumn{3}{c}{VQA Datasets} \\
& PMCPM & ROCO+CXR & PMCPM & ROCO+CXR & VQAMed19 & VQARAD & SLAKE \\
\midrule
(a) & \xmark & \xmark & \xmark & \xmark &  77.60 & 74.06 & 81.43\\
(b) & \checkmark & \xmark & \xmark & \xmark & 78.40 & 74.94 & 83.60\\
(c) & \xmark & \checkmark & \xmark & \xmark &  79.73 & 75.17 & 82.94 \\
(d) & \checkmark & \checkmark & \xmark & \xmark & 80.27 & 76.72 & 85.58 \\
\midrule
(e) & \xmark & \checkmark & \xmark & \checkmark & 80.00 & 76.72 & 85.20 \\
(f) & \checkmark & \checkmark  & \xmark & \checkmark & 81.07 & 77.16 & 85.67 \\
(g) & \checkmark & \checkmark  & \checkmark & \checkmark & \textbf{82.13} & \textbf{78.27} & \textbf{86.05}  \\
\bottomrule
\end{tabular}
\caption{Results of RAMM on the VQA-Med 2019, VQARAD, and SLAKE with different pretraining and retrieval settings.}
\label{tab:pretrain corpus}
\end{table*}

\begin{table*}[t]
\small
\centering
\begin{tabular}{lcccccc}
\toprule
Task & \multicolumn{2}{c}{VQAMed19} & \multicolumn{2}{c}{VQARAD} & \multicolumn{2}{c}{SLAKE} \\
Dataset & Retrieve\% & Have Ans\% & Retrieve\% & Have Ans\% & Retrieve\% & Have Ans\% \\
\midrule
PMCPM + ROCO + MIMIC-CXR &100.0\%&29.6\%&100.0\%&28.4\%& 100.0\%&38.7\%  \\
PMCPM &52.2\%&20.8\%&51.9\%&\textbf{21.3\%}& 54.9\%& \textbf{30.5\%} \\
 ROCO + MIMIC-CXR  & 47.8\% & \textbf{21.1\%} & 48.1\% & 20.0\% & 45.1\% & 26.9\% \\
\bottomrule
\end{tabular}
\caption{Analysis of the source and quality of retrieving on VQA-Med 2019, VQARAD, and SLAKE datasets. 
}
\label{tab:retrievalmain}
\end{table*}

\subsection{Ablation Study}

To further analyze the influence of the collected PMCPM dataset and the retrieval strategy, we conduct detailed ablation studies as shown in Table \ref{tab:pretrain corpus}. 

Firstly, we analyze the impact of using different data during the pre-training phase. ROCO and MIMIC-CXR are well-used biomedical multi-modal pre-training datasets. ROCO contains labeled radiology images in relatively small amounts. The amount of MIMIC-CXR is large, while it only contains chest X-Ray image-text pairs. The collection of PMCPM comes from the rich and diverse conditions of patients, and PMCPM is much larger than ROCO and MIMIC-CXR.
With the numerous and diverse image-text pairs from PMCPM, the model has the potential to obtain better visual and textual representations. Comparing (a) and (b), we show the effectiveness of pretraining on our collected PMCPM dataset. Results from (b) and (c) show that the data of PMCPM and previous ROCO plus MIMIC-CXR can complement each other. ROCO plus MIMIC-CXR shows better results on VQA-Med-2019 and VQARAD datasets, while PMCPM is better on SLAKE dataset. By combining the two settings with three resources, setting (d) achieves the best performance with a large margin by only using either of them. The above results demonstrate that PMCPM can enhance the model pretraining in biomedical VQA tasks.

Secondly, the above datasets are also treated as retrieval resources in this work. Therefore, we also analyze the impact of using different data during the fine-tuning phase. Setting (e) only uses the previous ROCO and MIMIC-CXR datasets for retrieval. By comparing it with setting (c), we show that the proposed retrieval strategy is effective in the previously widely-used resource. Then, we further demonstrate the usefulness of our collected PMCPM data. Compared with settings (f) and (g), we observe that retrieving additional PMCPM data yields gains of 1.06, 1.11, and 0.38, respectively. 
In addition, results from setting (g) with (d) reflect the great improvement brought by our proposed retrieval method. The above ablation studies show that our collected PMCPM dataset and our proposed retrieval method can benefit both pretraining and retrieving in biomedical VQA tasks.

\subsection{Analysis and Discussion}

\subsubsection{Retrieval}

\paragraph{Retrieval distribution}

To understand how RAMM retrieve from different resources and how retrieved texts help our model for biomedical VQA tasks, we count the sources of retrieved samples and how many retrieved texts contain the answers.
Statistics are listed in Table~\ref{tab:retrievalmain}.
There are 29.6\%, 28.4\%, and 38.7\% of answers that can be found in retrieved texts in VQA-Med 2019, VQARAD, and SLAKE respectively, which are considerable amounts.
These retrieved texts may help RAMM answer the questions directly.
We observe that in these tasks, RAMM retrieves image-text pairs from PMCPM over half which shows that PMCPM is more related to biomedical VQA images than ROCO and MIMIC-CXR.
The percentages of retrieved texts containing answers are nearly the same for PMCPM and \textit{out of PMCPM} in VQA-Med 2019 and VQARAD. In SLAKE, PMCPM contributes more texts containing answers.
To conclude, PMCPM is important during retrieval which supplies additional texts containing answers.


\begin{table}[t]
\small
\centering
\begin{tabular}{cccc}
\toprule
\# Retrieval & VQARAD & SLAKE \\
\midrule
0 &  76.72 & 85.58 \\
1 & 77.38 & \textbf{86.33} \\
2 & 78.05 & \textbf{86.33} \\
4 &  \textbf{78.27} & 86.05 \\
8 &  76.50 & 84.54 \\
\bottomrule
\end{tabular}
\caption{Results on the VQARAD and SLAKE datasets with different $r$ of retrieved image-text pairs.}
\label{tab:count}
\end{table}

\begin{table}[t]
\small
\centering
\begin{tabular}{lcccc}
\toprule
Model & \multicolumn{2}{c}{VQARAD} & \multicolumn{2}{c}{SLAKE} \\
      & Open & Closed & Open & Closed \\
\midrule 
MEVF-BAN & 49.20 & 77.20 & 77.80 & 79.80 \\
CPRD-BAN & 52.50 & 77.90 & 79.50 & 83.40 \\
M3AE & 67.23 & 83.46 & 80.31 & 87.82 \\
\midrule 
RAMM & \textbf{67.60} & \textbf{85.29} & \textbf{82.48} & \textbf{91.59} \\
\bottomrule 
\end{tabular}
\caption{Analysis of open-ended and closed-ended results on the VQARAD and SLAKE datasets.}
\label{tab:openclose}
\end{table}

\paragraph{Image-text pairs retrieval count} To explore how image-text pairs retrieval count influences biomedical VQA fine-tuning, we search the number of retrieved images among $\{0,1,2,4,8\}$. Results for VQARAD and SLAKE are shown in Table~\ref{tab:count}.
Compared with vanilla fine-tuning (i.e. retrieval count 0), retrieving $\{1,2,4\}$ image-text pairs can boost performances. 
We empirically find retrieving 2 or 4 image-text pairs work well.
Furthermore, using 8 image-text pairs downgrades the model performance, which indicates that retrieving too many image-text pairs may introduce unrelated image-text pairs with noisy knowledge.

\subsubsection{Question Type}
In biomedical VQA, questions can be divided into closed-ended and open-ended questions. 
The answers to closed-ended questions are limited to multiple choices like \textit{Yes} or \textit{No}, while the answers to open-ended questions are not limited.
Closed-ended questions are easier than open-ended questions due to the form.
Table~\ref{tab:openclose} lists comparisons among different methods on open-ended and closed-ended questions.
RAMM performs better than M3AE on both closed-ended and open-ended questions.
RAMM improves M3AE on closed-ended questions more than open-ended questions.
The reason is that closed-ended questions are limited in answer choices, and are easier to be covered by retrieved texts.

\subsubsection{Comparison with ROCO}
ROCO and PMCPM are both subsets of PubMed.
ROCO uses a trained neural network to filter radiology images, while PMCPM finds images by ``\textit{Case Report}'' from PubMed.
We calculate the involved article count in ROCO and PMCPM train split.
ROCO uses 40,857 articles from PubMed, and PMCPM uses 127,455 articles from PubMed. 
There are 17,189 articles appeared in both datasets.
Although ROCO and PMCPM filter image-text pairs using different methods, nearly half of the pairs in ROCO are also collected by PMCPM. 
Considering ROCO is a manually corrected dataset, it can reflect the quality of the larger dataset PMCPM from the side.

\subsubsection{Case Study}
\begin{figure}[t!]
    \centering
    \includegraphics[width=1\linewidth]{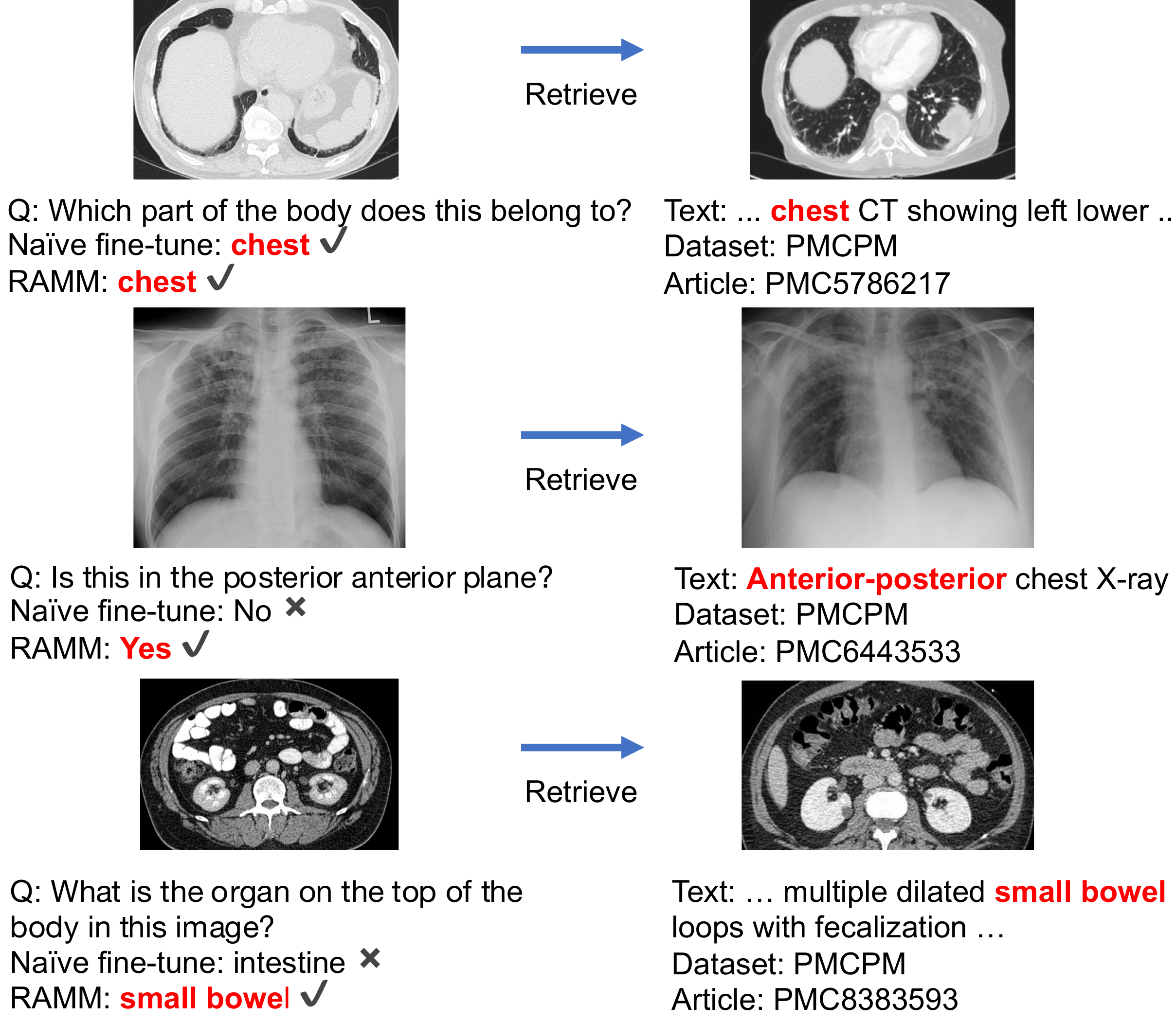}
    \caption{Examples of retrieved image-text pairs of RAMM. 
    }
    \label{fig:casestudy}
\end{figure}

We show retrieved image-text pairs of RAMM in Figure~\ref{fig:casestudy}.
The retrieved samples are all from PMCPM.
Retrieved images are in the same modalities and are the same organs as the original images.
These cases show that the retrieved images contain useful information for answering questions.
Although the answer (i.e. \textit{Yes}/\textit{No}) is not covered in retrieved text for the second example, the retrieved text is still useful for answering.
The second and third examples further show that RAMM can answer correctly with retrieval augmentation while naive fine-tuning fails.

\section{Conclusion}

In this paper, we present a retrieval-augmented pretrain-and-finetune paradigm RAMM for biomedical VQA which includes a high-quality image-text pairs PMCPM, a pretrained multi-modal model, and a novel retrieval-augmented attention module for fine-tuning.
RAMM uses ITC to retrieve related image-text pairs and fuse retrieved information with retrieval-attention.
Experiments have shown that our approach outperforms state-of-the-art methods in various biomedical VQA datasets.
Ablation studies show the effectiveness of proposed RAMM and PMCPM for pretraining and retrieving.

{\small
\bibliographystyle{ieee_fullname}
\bibliography{egbib}

\begin{thebibliography}{10}\itemsep=-1pt

\bibitem{beltagy-etal-2019-scibert}
Iz Beltagy, Kyle Lo, and Arman Cohan.
\newblock {S}ci{BERT}: A pretrained language model for scientific text.
\newblock In {\em Proceedings of the 2019 Conference on Empirical Methods in
  Natural Language Processing and the 9th International Joint Conference on
  Natural Language Processing (EMNLP-IJCNLP)}, pages 3615--3620, Hong Kong,
  China, Nov. 2019. Association for Computational Linguistics.

\bibitem{medvqa2019}
Asma Ben~Abacha, Sadid~A Hasan, Vivek~V Datla, Dina Demner-Fushman, and Henning
  M{\"u}ller.
\newblock Vqa-med: Overview of the medical visual question answering task at
  imageclef 2019.
\newblock In {\em Proceedings of CLEF (Conference and Labs of the Evaluation
  Forum) 2019 Working Notes}. 9-12 September 2019, 2019.

\bibitem{mevf}
D.~Nguyen Binh, Do Thanh-Toan, X.~Nguyen Binh, Do Tuong, Tjiputra Erman, and
  D.~Tran Quang.
\newblock Overcoming data limitation in medical visual question answering.
\newblock In {\em MICCAI}, 2019.

\bibitem{chen2022m3ae}
Zhihong Chen, Yuhao Du, Jinpeng Hu, Yang Liu, Guanbin Li, Xiang Wan, and
  Tsung-Hui Chang.
\newblock Multi-modal masked autoencoders for medical vision-and-language
  pre-training.
\newblock In {\em International Conference on Medical Image Computing and
  Computer-Assisted Intervention}. Springer, 2022.

\bibitem{arl}
Zhihong Chen, Guanbin Li, and Xiang Wan.
\newblock Align, reason and learn: Enhancing medical vision-and-language
  pre-training with knowledge.
\newblock In {\em Proceedings of the 30th ACM International Conference on
  Multimedia}, pages 5152--5161, 2022.

\bibitem{randaugment}
Ekin~Dogus Cubuk, Barret Zoph, Jon Shlens, and Quoc Le.
\newblock Randaugment: Practical automated data augmentation with a reduced
  search space.
\newblock In H. Larochelle, M. Ranzato, R. Hadsell, M.F. Balcan, and H. Lin,
  editors, {\em Advances in Neural Information Processing Systems}, volume~33,
  pages 18613--18624. Curran Associates, Inc., 2020.

\bibitem{delbrouck-etal-2022-vilmedic}
Jean-benoit Delbrouck, Khaled Saab, Maya Varma, Sabri Eyuboglu, Pierre Chambon,
  Jared Dunnmon, Juan Zambrano, Akshay Chaudhari, and Curtis Langlotz.
\newblock {V}i{LM}edic: a framework for research at the intersection of vision
  and language in medical {AI}.
\newblock In {\em Proceedings of the 60th Annual Meeting of the Association for
  Computational Linguistics: System Demonstrations}, pages 23--34, Dublin,
  Ireland, May 2022. Association for Computational Linguistics.

\bibitem{bert}
Jacob Devlin, Ming-Wei Chang, Kenton Lee, and Kristina Toutanova.
\newblock {BERT}: Pre-training of deep bidirectional transformers for language
  understanding.
\newblock In {\em Proceedings of the 2019 Conference of the North {A}merican
  Chapter of the Association for Computational Linguistics: Human Language
  Technologies, Volume 1 (Long and Short Papers)}, pages 4171--4186,
  Minneapolis, Minnesota, June 2019. Association for Computational Linguistics.

\bibitem{aioz_mmq_miccai21}
Tuong Do, Binh~X. Nguyen, Erman Tjiputra, Minh Tran, Quang~D. Tran, and Anh
  Nguyen.
\newblock Multiple meta-model quantifying for medical visual question
  answering.
\newblock In {\em MICCAI}, 2021.

\bibitem{dou2022meter}
Zi-Yi Dou, Yichong Xu, Zhe Gan, Jianfeng Wang, Shuohang Wang, Lijuan Wang,
  Chenguang Zhu, Pengchuan Zhang, Lu Yuan, Nanyun Peng, Zicheng Liu, and
  Michael Zeng.
\newblock An empirical study of training end-to-end vision-and-language
  transformers.
\newblock In {\em Conference on Computer Vision and Pattern Recognition
  (CVPR)}, 2022.

\bibitem{pubmedclip}
Sedigheh {Eslami}, Gerard {de Melo}, and Christoph {Meinel}.
\newblock Does {CLIP} benefit visual question answering in the medical domain
  as much as it does in the general domain?
\newblock {\em arXiv e-prints}, Dec. 2021.

\bibitem{mtpt}
Haifan Gong, Guanqi Chen, Sishuo Liu, Yizhou Yu, and Guanbin Li.
\newblock Cross-modal self-attention with multi-task pre-training for medical
  visual question answering.
\newblock In {\em {ICMR} '21: International Conference on Multimedia Retrieval,
  Taipei, Taiwan, August 21-24, 2021}, pages 456--460. {ACM}, 2021.

\bibitem{gong2021sysu-hcp}
Haifan Gong, Ricong Huang, Guanqi Chen, and Guanbin Li.
\newblock Sysu-hcp at vqa-med 2021: A data-centric model with efficient
  training methodology for medical visual question answering.
\newblock In {\em CLEF 2021 -- Conference and Labs of the Evaluation Forum,
  September 21--24, 2021, Bucharest, Romania}, CEUR Workshop Proceedings, 2021.

\bibitem{pubmedbert}
Yuxian Gu, Robert Tinn, Hao Cheng, Michael~R. Lucas, Naoto Usuyama, Xiaodong
  Liu, Tristan Naumann, Jianfeng Gao, and Hoifung Poon.
\newblock Domain-specific language model pretraining for biomedical natural
  language processing.
\newblock {\em ACM Transactions on Computing for Healthcare (HEALTH)}, 3:1 --
  23, 2022.

\bibitem{guu2020retrieval}
Kelvin Guu, Kenton Lee, Zora Tung, Panupong Pasupat, and Mingwei Chang.
\newblock Retrieval augmented language model pre-training.
\newblock In {\em International Conference on Machine Learning}, pages
  3929--3938. PMLR, 2020.

\bibitem{medvqa2021}
Bogdan Ionescu, Henning M{\"u}ller, Renaud P{\'e}teri, Asma~Ben Abacha, Mourad
  Sarrouti, Dina Demner-Fushman, Sadid~A Hasan, Serge Kozlovski, Vitali
  Liauchuk, Yashin~Dicente Cid, et~al.
\newblock Overview of the imageclef 2021: Multimedia retrieval in medical,
  nature, internet and social media applications.
\newblock In {\em International Conference of the Cross-Language Evaluation
  Forum for European Languages}, pages 345--370. Springer, 2021.

\bibitem{bioqa_review}
Qiao Jin, Zheng Yuan, Guangzhi Xiong, Qianlan Yu, Huaiyuan Ying, Chuanqi Tan,
  Mosha Chen, Songfang Huang, Xiaozhong Liu, and Sheng Yu.
\newblock Biomedical question answering: A survey of approaches and challenges.
\newblock {\em ACM Comput. Surv.}, 55(2), jan 2022.

\bibitem{mimic-cxr}
Alistair E.~W. Johnson, Tom~J. Pollard, Seth~J. Berkowitz, Nathaniel~R.
  Greenbaum, Matthew~P. Lungren, Chih ying Deng, Roger~G. Mark, and Steven
  Horng.
\newblock Mimic-cxr, a de-identified publicly available database of chest
  radiographs with free-text reports.
\newblock {\em Scientific Data}, 6, 2019.

\bibitem{faiss}
Jeff Johnson, Matthijs Douze, and Herv{\'e} J{\'e}gou.
\newblock Billion-scale similarity search with {GPUs}.
\newblock {\em IEEE Transactions on Big Data}, 7(3):535--547, 2019.

\bibitem{khare2021mmbert}
Yash Khare, Viraj Bagal, Minesh Mathew, Adithi Devi, U~Deva Priyakumar, and CV
  Jawahar.
\newblock Mmbert: multimodal bert pretraining for improved medical vqa.
\newblock In {\em 2021 IEEE 18th International Symposium on Biomedical Imaging
  (ISBI)}, pages 1033--1036. IEEE, 2021.

\bibitem{ban}
Jin-Hwa Kim, Jaehyun Jun, and Byoung-Tak Zhang.
\newblock {Bilinear Attention Networks}.
\newblock In {\em Advances in Neural Information Processing Systems 31}, pages
  1571--1581, 2018.

\bibitem{vqarad}
Jason~J Lau, Soumya Gayen, Asma Ben~Abacha, and Dina Demner-Fushman.
\newblock A dataset of clinically generated visual questions and answers about
  radiology images.
\newblock {\em Scientific data}, 5(1):1--10, 2018.

\bibitem{lee2020biobert}
Jinhyuk Lee, Wonjin Yoon, Sungdong Kim, Donghyeon Kim, Sunkyu Kim, Chan~Ho So,
  and Jaewoo Kang.
\newblock Biobert: a pre-trained biomedical language representation model for
  biomedical text mining.
\newblock {\em Bioinformatics}, 36(4):1234--1240, 2020.

\bibitem{lee2019latent}
Kenton Lee, Ming-Wei Chang, and Kristina Toutanova.
\newblock Latent retrieval for weakly supervised open domain question
  answering.
\newblock In {\em Proceedings of the 57th Annual Meeting of the Association for
  Computational Linguistics}, pages 6086--6096, Florence, Italy, July 2019.
  Association for Computational Linguistics.

\bibitem{lewis2020retrieval}
Patrick Lewis, Ethan Perez, Aleksandra Piktus, Fabio Petroni, Vladimir
  Karpukhin, Naman Goyal, Heinrich K{\"u}ttler, Mike Lewis, Wen-tau Yih, Tim
  Rockt{\"a}schel, et~al.
\newblock Retrieval-augmented generation for knowledge-intensive nlp tasks.
\newblock {\em Advances in Neural Information Processing Systems},
  33:9459--9474, 2020.

\bibitem{li2022mplug}
Chenliang Li, Haiyang Xu, Junfeng Tian, Wei Wang, Ming Yan, Bin Bi, Jiabo Ye,
  Hehong Chen, Guohai Xu, Zheng Cao, et~al.
\newblock mplug: Effective and efficient vision-language learning by
  cross-modal skip-connections.
\newblock {\em arXiv preprint arXiv:2205.12005}, 2022.

\bibitem{albef}
Junnan Li, Ramprasaath~R. Selvaraju, Akhilesh~Deepak Gotmare, Shafiq Joty,
  Caiming Xiong, and Steven Hoi.
\newblock Align before fuse: Vision and language representation learning with
  momentum distillation.
\newblock In {\em NeurIPS}, 2021.

\bibitem{liang2021rdrop}
Xiaobo Liang, Lijun Wu, Juntao Li, Yue Wang, Qi Meng, Tao Qin, Wei Chen, Min
  Zhang, and Tie-Yan Liu.
\newblock R-drop: Regularized dropout for neural networks.
\newblock In {\em NeurIPS}, 2021.

\bibitem{cprd}
Bo Liu, Li-Ming Zhan, and Xiao-Ming Wu.
\newblock Contrastive pre-training and representation distillation for medical
  visual question answering based on radiology images.
\newblock In {\em International Conference on Medical Image Computing and
  Computer-Assisted Intervention}, pages 210--220. Springer, 2021.

\bibitem{slake}
Bo Liu, Li-Ming Zhan, Li Xu, Lin Ma, Yan Yang, and Xiao-Ming Wu.
\newblock Slake: a semantically-labeled knowledge-enhanced dataset for medical
  visual question answering.
\newblock In {\em 2021 IEEE 18th International Symposium on Biomedical Imaging
  (ISBI)}, pages 1650--1654. IEEE, 2021.

\bibitem{swin}
Ze Liu, Yutong Lin, Yue Cao, Han Hu, Yixuan Wei, Zheng Zhang, Stephen Lin, and
  Baining Guo.
\newblock Swin transformer: Hierarchical vision transformer using shifted
  windows.
\newblock In {\em Proceedings of the IEEE/CVF International Conference on
  Computer Vision}, pages 10012--10022, 2021.

\bibitem{long2022retrieval}
Alexander Long, Wei Yin, Thalaiyasingam Ajanthan, Vu Nguyen, Pulak Purkait,
  Ravi Garg, Alan Blair, Chunhua Shen, and Anton van~den Hengel.
\newblock Retrieval augmented classification for long-tail visual recognition.
\newblock In {\em Proceedings of the IEEE/CVF Conference on Computer Vision and
  Pattern Recognition}, pages 6959--6969, 2022.

\bibitem{loshchilov2017decoupled}
Ilya Loshchilov and Frank Hutter.
\newblock Decoupled weight decay regularization.
\newblock In {\em 7th International Conference on Learning Representations,
  {ICLR} 2019, New Orleans, LA, USA, May 6-9, 2019}, 2019.

\bibitem{medvill}
Jong~Hak Moon, Hyungyung Lee, Woncheol Shin, Young-Hak Kim, and Edward Choi.
\newblock Multi-modal understanding and generation for medical images and text
  via vision-language pre-training.
\newblock {\em IEEE Journal of Biomedical and Health Informatics}, 2022.

\bibitem{nguyen2019overcoming}
Binh~D Nguyen, Thanh-Toan Do, Binh~X Nguyen, Tuong Do, Erman Tjiputra, and
  Quang~D Tran.
\newblock Overcoming data limitation in medical visual question answering.
\newblock In {\em International Conference on Medical Image Computing and
  Computer-Assisted Intervention}, pages 522--530. Springer, 2019.

\bibitem{roco}
Obioma Pelka, Sven Koitka, Johannes R{\"u}ckert, Felix Nensa, and Christoph~M
  Friedrich.
\newblock Radiology objects in context (roco): a multimodal image dataset.
\newblock In {\em Intravascular Imaging and Computer Assisted Stenting and
  Large-Scale Annotation of Biomedical Data and Expert Label Synthesis}, pages
  180--189. Springer, 2018.

\bibitem{clip}
Alec Radford, Jong~Wook Kim, Chris Hallacy, Aditya Ramesh, Gabriel Goh,
  Sandhini Agarwal, Girish Sastry, Amanda Askell, Pamela Mishkin, Jack Clark,
  et~al.
\newblock Learning transferable visual models from natural language
  supervision.
\newblock In {\em International Conference on Machine Learning}, pages
  8748--8763. PMLR, 2021.

\bibitem{bm25}
Stephen~E. Robertson and Hugo Zaragoza.
\newblock The probabilistic relevance framework: Bm25 and beyond.
\newblock {\em Found. Trends Inf. Retr.}, 3:333--389, 2009.

\bibitem{sarto2022retrieval}
Sara Sarto, Marcella Cornia, Lorenzo Baraldi, and Rita Cucchiara.
\newblock Retrieval-augmented transformer for image captioning.
\newblock In {\em International Conference on Content-based Multimedia
  Indexing}, pages 1--7, 2022.

\bibitem{vaswani2017attention}
Ashish Vaswani, Noam Shazeer, Niki Parmar, Jakob Uszkoreit, Llion Jones,
  Aidan~N Gomez, {\L}ukasz Kaiser, and Illia Polosukhin.
\newblock Attention is all you need.
\newblock {\em Advances in neural information processing systems}, 30, 2017.

\bibitem{wang2022image}
Wenhui Wang, Hangbo Bao, Li Dong, Johan Bjorck, Zhiliang Peng, Qiang Liu, Kriti
  Aggarwal, Owais~Khan Mohammed, Saksham Singhal, Subhojit Som, et~al.
\newblock Image as a foreign language: Beit pretraining for all vision and
  vision-language tasks.
\newblock {\em arXiv preprint arXiv:2208.10442}, 2022.

\bibitem{san}
Zichao Yang, Xiaodong He, Jianfeng Gao, Li Deng, and Alex Smola.
\newblock Stacked attention networks for image question answering.
\newblock In {\em Proceedings of the IEEE conference on computer vision and
  pattern recognition}, pages 21--29, 2016.

\bibitem{mfb}
Zhou Yu, Jun Yu, Jianping Fan, and Dacheng Tao.
\newblock Multi-modal factorized bilinear pooling with co-attention learning
  for visual question answering.
\newblock {\em IEEE International Conference on Computer Vision (ICCV)}, pages
  1839--1848, 2017.

\bibitem{yuan-etal-2022-biobart}
Hongyi Yuan, Zheng Yuan, Ruyi Gan, Jiaxing Zhang, Yutao Xie, and Sheng Yu.
\newblock {B}io{BART}: Pretraining and evaluation of a biomedical generative
  language model.
\newblock In {\em Proceedings of the 21st Workshop on Biomedical Language
  Processing}, pages 97--109, Dublin, Ireland, May 2022. Association for
  Computational Linguistics.

\bibitem{yuan-etal-2021-improving}
Zheng Yuan, Yijia Liu, Chuanqi Tan, Songfang Huang, and Fei Huang.
\newblock Improving biomedical pretrained language models with knowledge.
\newblock In {\em Proceedings of the 20th Workshop on Biomedical Language
  Processing}, pages 180--190, Online, June 2021. Association for Computational
  Linguistics.

\bibitem{condr}
Li-Ming Zhan, Bo Liu, Lu Fan, Jiaxin Chen, and Xiao-Ming Wu.
\newblock Medical visual question answering via conditional reasoning.
\newblock In {\em Proceedings of the 28th ACM International Conference on
  Multimedia}, pages 2345--2354, 2020.

\bibitem{zhao2022pmc}
Zhengyun Zhao, Qiao Jin, and Sheng Yu.
\newblock Pmc-patients: A large-scale dataset of patient notes and relations
  extracted from case reports in pubmed central.
\newblock {\em arXiv preprint arXiv:2202.13876}, 2022.

\end{thebibliography}
}

\end{document}